\documentclass[conference]{IEEEtran}

\usepackage{blindtext}
\usepackage{eso-pic}
\IEEEoverridecommandlockouts
\usepackage{cite}
\usepackage{amsmath,amssymb,amsfonts}
\usepackage{algorithmic}
\usepackage{graphicx}
\usepackage{textcomp}
\usepackage{xcolor}
\usepackage{booktabs}
\usepackage{multirow}
\usepackage{hyperref}

\def\BibTeX{{\rm B\kern-.05em{\sc i\kern-.025em b}\kern-.08em
    T\kern-.1667em\lower.7ex\hbox{E}\kern-.125emX}}

\usepackage{eso-pic}

\begin{document}

\title{\vspace*{1cm} Enhancing Financial Report Question-Answering: A Retrieval-Augmented Generation System with Reranking Analysis%
\thanks{Accepted at the 6th International Conference on Electrical, Computer and Energy Technologies (ICECET 2026), Rome, Italy, July 2026.}
}

\author{\IEEEauthorblockN{1\textsuperscript{st} Zhiyuan Cheng}
\IEEEauthorblockA{\textit{School of Engineering} \\
\textit{Stanford University}\\
Stanford, CA, USA \\
zhycheng@stanford.edu}
\and
\IEEEauthorblockN{2\textsuperscript{nd} Longying Lai}
\IEEEauthorblockA{\textit{Simon Business School} \\
\textit{University of Rochester}\\
Rochester, NY, USA \\
longyinglai703@gmail.com}
\and
\IEEEauthorblockN{3\textsuperscript{rd} Yue Liu}
\IEEEauthorblockA{\textit{Accounting \& Information Systems} \\
\textit{Rutgers University}\\
Newark, NJ, USA \\
yl1641@scarletmail.rutgers.edu}
\and
\IEEEauthorblockN{4\textsuperscript{th} Kai Cheng}
\IEEEauthorblockA{\textit{Institute for Social and Economic Research and Policy} \\
\textit{Columbia University}\\
New York, NY, USA \\
kc2859@columbia.edu}
\and
\IEEEauthorblockN{5\textsuperscript{th} Xiaoxi Qi}
\IEEEauthorblockA{\textit{Department of Economics} \\
\textit{Northeastern University}\\
Boston, MA, USA \\
xiaoxiq15@gmail.com}
}

\maketitle

\begin{abstract}
Financial analysts face significant challenges extracting information from lengthy 10-K reports, which often exceed 100 pages. This paper presents a Retrieval-Augmented Generation (RAG) system designed to answer questions about S\&P 500 financial reports and evaluates the impact of neural reranking on system performance. Our pipeline employs hybrid search combining full-text and semantic retrieval, followed by an optional reranking stage using a cross-encoder model. We conduct systematic evaluation using the FinDER benchmark dataset, comprising 1,500 queries across five experimental groups. Results demonstrate that reranking significantly improves answer quality, achieving 49.0 percent correctness for scores of 8 or above compared to 33.5 percent without reranking, representing a 15.5 percentage point improvement. Additionally, the error rate for completely incorrect answers decreases from 35.3 percent to 22.5 percent. Our findings emphasize the critical role of reranking in financial RAG systems and demonstrate performance improvements over baseline methods through modern language models and refined retrieval strategies.
\end{abstract}


\section{Introduction}

\subsection{Background and Motivation}

Financial 10-K reports represent comprehensive annual filings required by the U.S. Securities and Exchange Commission (SEC) for publicly traded companies. These documents typically span 100-300 pages and contain detailed information about a company's financial performance, operations, risk factors, and management discussions. Financial analysts, investors, and researchers routinely consult these reports to make informed investment decisions, assess company performance, and understand business strategies.

However, the sheer volume and complexity of these documents pose significant challenges for manual information extraction. Analysts must navigate dense prose, financial tables, and regulatory disclosures to locate specific information relevant to their queries. This process is time-consuming, error-prone, and does not scale efficiently across multiple companies or time periods.

Retrieval-Augmented Generation (RAG) \cite{lewis2020rag} offers a promising solution to this challenge by combining information retrieval with large language model (LLM) generation capabilities. Unlike pure generative approaches that rely solely on parametric knowledge, RAG systems retrieve relevant document passages to ground their responses, reducing hallucinations and improving factual accuracy. This approach is particularly valuable in the financial domain, where precision is paramount and errors can lead to significant financial consequences.

\subsection{Problem Statement}

Despite the promise of RAG systems, achieving high accuracy in financial question-answering remains challenging. The quality of retrieved information directly impacts the final answer quality—irrelevant or marginally relevant passages can mislead the generation model, resulting in incorrect or incomplete responses. Traditional retrieval methods, including keyword-based search and semantic similarity search, often struggle with the nuanced language and domain-specific terminology prevalent in financial documents.

This paper addresses a critical research question: \textbf{How much does neural reranking contribute to RAG performance in the financial domain?} While reranking has shown promise in general-domain information retrieval, its impact on financial document question-answering, particularly with modern LLMs, remains underexplored.

\subsection{Contributions}

This work makes the following contributions:

\begin{enumerate}
\item \textbf{Complete RAG Pipeline Implementation:} We present an end-to-end RAG system for financial report question-answering, processing all 500 S\&P 500 companies' 10-K reports with a hybrid retrieval approach combining full-text search and semantic similarity.

\item \textbf{Systematic Benchmark Evaluation:} We conduct rigorous evaluation using 1,500 queries from the FinDER dataset \cite{finder2025}, organized into five independent groups of 300 queries each to ensure statistical robustness.

\item \textbf{Empirical Reranking Analysis:} We provide empirical evidence that neural reranking significantly improves performance, achieving a 15.5 percentage point improvement in correctness rate (49.0\% vs. 33.5\%) and reducing completely incorrect answers by 12.8 percentage points.

\item \textbf{Performance Improvement Demonstration:} Our system achieves approximately 49\% correctness on the FinDER benchmark, representing substantial improvement over baseline performance reported in prior work ($\sim$33\% correctness), attributable to modern LLMs (GPT-4.1), neural reranking, and hybrid retrieval strategies.
\end{enumerate}

\subsection{Paper Organization}

The remainder of this paper is organized as follows: Section II reviews related work on RAG systems, financial document analysis, and retrieval enhancement techniques. Section III describes our system architecture in detail. Section IV presents the experimental setup, including dataset description and evaluation methodology. Section V discusses results and provides analysis. Section VI concludes with future research directions.

\section{Related Work}

\subsection{Retrieval-Augmented Generation}

Retrieval-Augmented Generation, introduced by Lewis et al. \cite{lewis2020rag}, combines parametric memory (pre-trained language models) with non-parametric memory (external document collections) to improve factual accuracy in knowledge-intensive NLP tasks. The RAG paradigm has since been applied to various domains, including open-domain question answering \cite{lewis2020rag}, dialogue systems, and domain-specific applications.

The core RAG architecture consists of two components: a retriever that identifies relevant documents given a query, and a generator that produces natural language responses conditioned on both the query and retrieved documents. This approach addresses limitations of pure generative models, which struggle to access precise knowledge, cannot provide provenance for their responses, and cannot easily update their world knowledge \cite{lewis2020rag}.

\subsection{Financial Document Analysis}

Financial document analysis has received increasing attention from the NLP community. The FinDER dataset \cite{finder2025}, introduced by LinqAlpha, provides 5,703 query-evidence-answer triplets derived from real-world financial inquiries across 10-K filings. Unlike existing financial QA datasets, FinDER reflects realistic analyst workflows with ambiguous, concise queries featuring domain-specific abbreviations and jargon.

The dataset's characteristics present unique challenges: 84.5\% qualitative questions, 15.5\% quantitative questions (with half requiring multi-step reasoning), and relevant evidence often scattered across multiple document sections \cite{finder2025}. Multi-hop reasoning in RAG systems has been explored through graph-based and tree-based retrieval approaches \cite{shi2026reasoning}. Prior work reports that advanced models like GPT-4-Turbo achieve only 9\% accuracy on closed-book financial questions, highlighting the necessity of retrieval-augmented approaches \cite{finder2025}.

\subsection{Retrieval Enhancement Techniques}

Modern retrieval systems employ various techniques to improve result quality:

\textbf{Hybrid Search:} Combining traditional keyword-based methods (BM25, full-text search) with dense semantic retrieval addresses complementary strengths—keyword search excels at exact matching while semantic search captures conceptual similarity \cite{robertson2009bm25}.

\textbf{Reciprocal Rank Fusion (RRF):} Cormack et al. \cite{cormack2009rrf} introduced RRF as an unsupervised method for combining multiple ranked lists. Using the formula $\text{RRFscore}(d) = \sum \frac{1}{k + r(d)}$, where $k=60$ and $r(d)$ is the rank of document $d$, RRF consistently outperforms individual systems and other meta-ranking methods without requiring training data \cite{cormack2009rrf}.

\textbf{Neural Reranking:} Cross-encoder models re-score retrieved candidates by jointly encoding query-document pairs, capturing fine-grained relevance signals missed by first-stage retrievers \cite{reimers2019sbert}. Recent models like Jina Reranker v2 \cite{jina2024reranker} employ flash attention mechanisms and support multilingual contexts, achieving state-of-the-art performance on benchmark datasets. Context pruning techniques \cite{fang2025attentionrag} further refine retrieved context by selecting the most relevant passages, complementing reranking approaches.

\subsection{Gap Analysis}

While reranking has demonstrated effectiveness in general information retrieval, its impact on financial RAG systems remains underexplored, particularly with modern large language models. Most existing studies focus on individual components (retrieval or generation) rather than evaluating the end-to-end pipeline. Additionally, systematic evaluation with realistic financial queries and proper statistical controls is limited. Our work addresses these gaps through controlled experimentation on a standardized benchmark.

\section{System Architecture}

\begin{figure}[!t]
\centering
\includegraphics[width=\columnwidth]{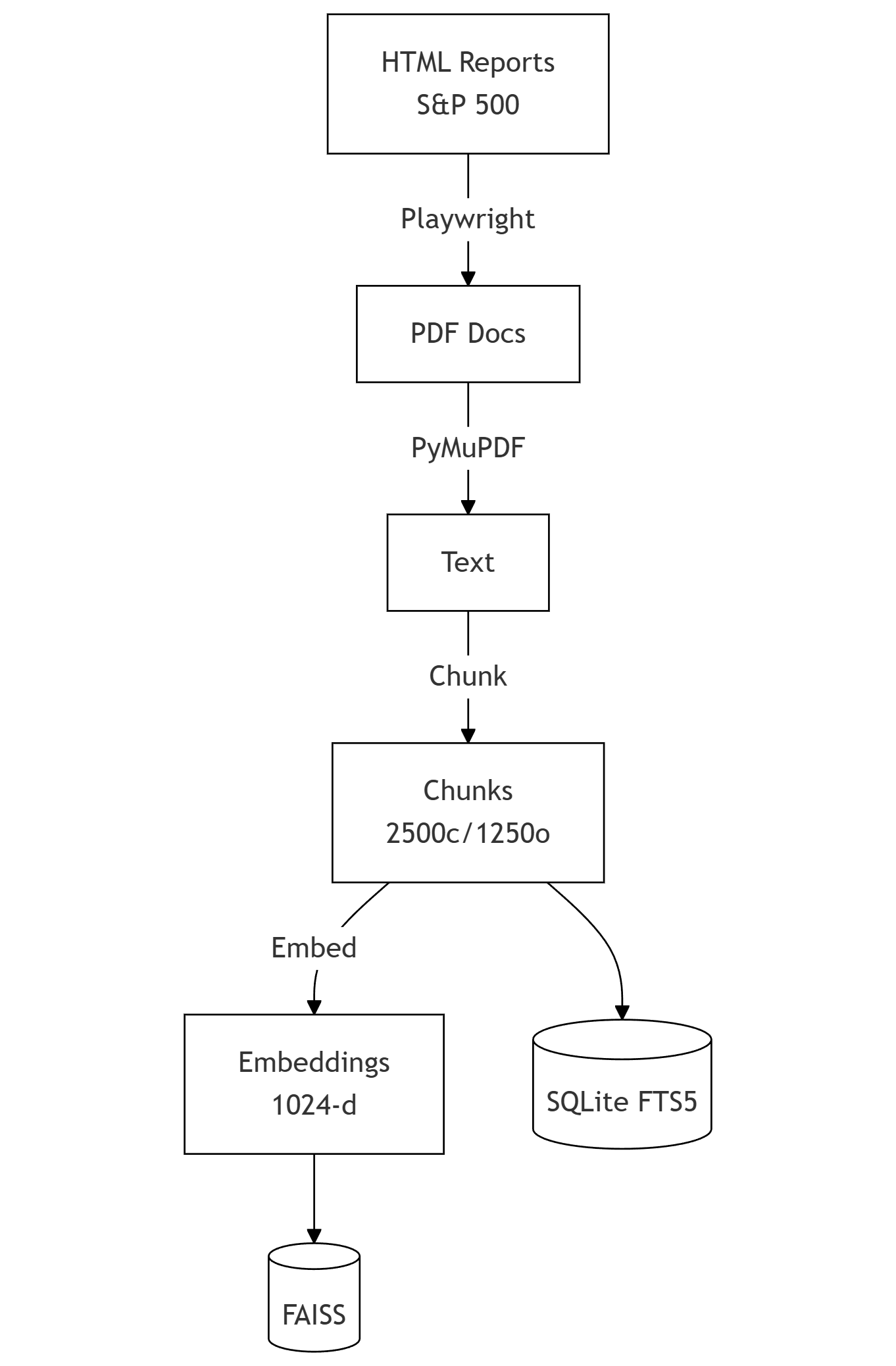}
\caption{Document Processing Pipeline. The system converts HTML reports to PDF, extracts and chunks text, generates embeddings, and stores data in SQLite (for keyword search) and FAISS (for semantic search).}
\label{fig:doc_pipeline}
\end{figure}

\begin{figure}[!htbp]
\centering
\includegraphics[width=\columnwidth]{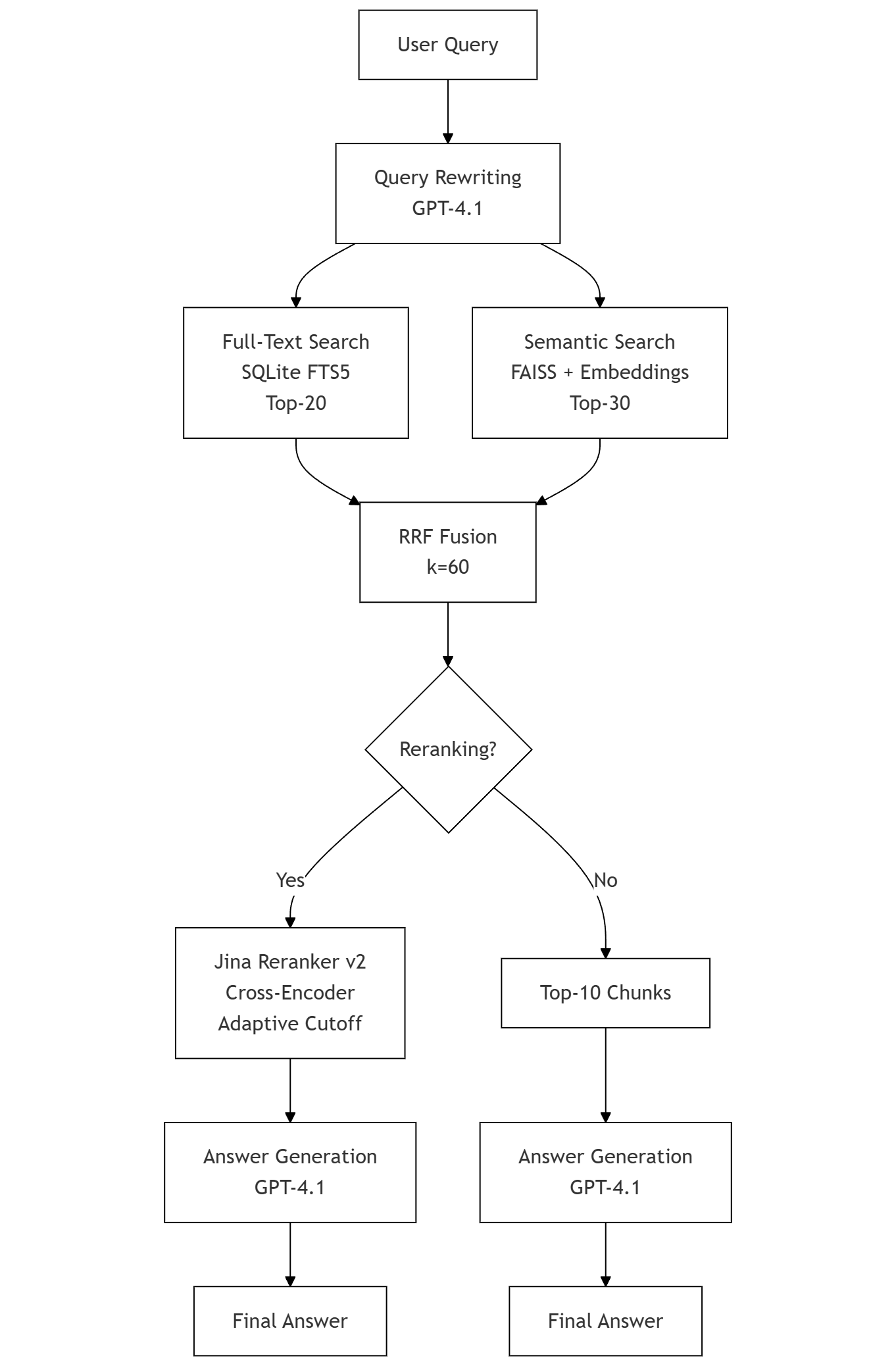}
\caption{Query Processing Pipeline with Reranking Ablation. The pipeline processes queries through rewriting, hybrid search (FTS + semantic), RRF fusion, optional reranking, and answer generation. The ablation study compares performance with and without the reranking stage.}
\label{fig:query_pipeline}
\end{figure}

Our RAG pipeline consists of two main stages: document processing (offline) and query processing (online). Fig. \ref{fig:doc_pipeline} illustrates the document processing flow, and Fig. \ref{fig:query_pipeline} shows the query processing architecture.

\subsection{Document Processing Pipeline}

\subsubsection{Data Collection}

We process 10-K reports from all 500 S\&P 500 companies. Documents are obtained in HTML format from company websites and converted to PDF using Playwright \cite{playwright2024}, a browser automation tool that renders HTML with proper formatting and layout preservation. This conversion step ensures consistent text extraction across documents.

\subsubsection{Text Extraction and Chunking}

PDF documents are parsed using PyMuPDF to extract textual content while preserving document structure. We employ fixed-size chunking with the following parameters:

\begin{itemize}
\item \textbf{Chunk size:} 2,500 characters
\item \textbf{Chunk overlap:} 1,250 characters (50\% overlap)
\end{itemize}

The overlap strategy ensures that information spanning chunk boundaries is not lost. These parameters balance context preservation (larger chunks retain more context) with retrieval granularity (smaller chunks enable more precise matching).

Extracted chunks are stored in a SQLite database with FTS5 (Full-Text Search) extension \cite{sqlite2024fts5}, enabling efficient keyword-based retrieval with support for Boolean queries, phrase matching, and relevance ranking.

\subsubsection{Embedding Generation}

Each text chunk is encoded into a dense vector representation using OpenAI's text-embedding-3-small model \cite{openai2024embeddings}, which generates 1,024-dimensional embeddings. Embeddings are computed in batches of 32 for efficiency. Empirical comparisons of embedding methods have shown that classical word embeddings can outperform transformer-based models for very short industrial texts, while transformer architectures demonstrate stronger performance on longer domain-specific passages \cite{lai2026transformers}. Our chunk size of 2,500 characters falls well within the range where transformer-based embeddings provide reliable semantic representations.

Generated embeddings are indexed using FAISS (Facebook AI Similarity Search) \cite{johnson2021faiss}, a library optimized for efficient similarity search and clustering of dense vectors. FAISS enables approximate nearest neighbor search with sub-linear time complexity, making it suitable for large-scale document collections.

\subsection{Query Processing Pipeline}

\subsubsection{Query Rewriting}

Raw user queries often contain ambiguity, typos, or lack specificity. We employ LLM-based query rewriting using GPT-4.1 to:

\begin{itemize}
\item Clarify vague terms with more specific language
\item Correct grammatical errors and typos
\item Extract keywords for full-text search
\end{itemize}

The rewriting module outputs two components: (1) a clarified query for semantic search, and (2) a list of extracted keywords for keyword-based retrieval. This preprocessing step improves retrieval recall by addressing query formulation issues.

\subsubsection{Hybrid Retrieval}

We implement a two-stage hybrid retrieval approach:

\textbf{Full-Text Search (FTS):} Using the extracted keywords, we query the SQLite FTS5 index to retrieve the top-20 most relevant chunks based on keyword matching. FTS excels at finding exact matches for terminology and named entities common in financial documents.

\textbf{Semantic Search:} The clarified query is embedded using the same text-embedding-3-small model, and we perform approximate nearest neighbor search in the FAISS index to retrieve the top-30 semantically similar chunks. We apply a distance threshold of 2.0 to filter out low-quality matches.

\textbf{Result Fusion:} Retrieved candidates from both methods are combined using Reciprocal Rank Fusion (RRF) \cite{cormack2009rrf} with $k=60$. RRF assigns each chunk a score based on its ranks in both retrieval lists:

\begin{equation}
\text{RRFscore}(d) = \frac{1}{60 + r_{\text{FTS}}(d)} + \frac{1}{60 + r_{\text{semantic}}(d)}
\end{equation}

This unsupervised fusion strategy leverages complementary strengths of both retrieval methods without requiring training data.

\subsubsection{Neural Reranking}

The reranking stage refines the candidate list from hybrid retrieval:

\textbf{Model:} We employ Jina Reranker v2 (jina-reranker-v2-base-multilingual) \cite{jina2024reranker}, a 278M-parameter cross-encoder model supporting 108 languages.

\textbf{Input:} Up to 30 chunks from the RRF-fused candidate list.

\textbf{Scoring:} The cross-encoder jointly encodes the query and each candidate chunk, computing fine-grained relevance scores that capture semantic similarity beyond first-stage retrievers.

\textbf{Cutoff Strategies:} We employ two adaptive cutoff mechanisms to select the most relevant chunks:

\begin{itemize}
\item \textit{Cumulative Probability Threshold:} Keep top chunks until cumulative probability mass reaches 55\% (i.e., stop when remaining probability $< 45\%$).
\item \textit{Score Cliff Detection:} If the score drops by more than 0.15 from the top-scoring chunk, truncate at that point to exclude low-quality candidates.
\end{itemize}

These strategies prevent dilution of context quality by filtering out marginally relevant chunks. The reranking stage typically produces 5-10 high-quality chunks.

\subsubsection{Answer Generation}

Retrieved chunks (top-10 without reranking, or reranked chunks with reranking) are assembled into a context string and provided to GPT-4.1 for answer generation. The prompt structure consists of:

\begin{itemize}
\item \textbf{System Prompt:} Instructs the model to answer based solely on provided context, avoiding speculation when information is insufficient.
\item \textbf{User Prompt:} Includes the retrieved context and the original user query.
\end{itemize}

We use deterministic sampling (temperature=0.0) during evaluation to ensure reproducibility.

\section{Experimental Setup}

\subsection{Dataset}

We evaluate our system using the FinDER benchmark \cite{finder2025}, a dataset specifically designed for financial question-answering and RAG evaluation.

\textbf{Dataset Characteristics:}
\begin{itemize}
\item Total size: 5,703 query-answer pairs
\item Domain: 10-K financial reports
\item Question types: 84.5\% qualitative, 15.5\% quantitative
\item Complexity: Multi-hop reasoning, domain-specific jargon, scattered evidence
\end{itemize}

\textbf{Sampling Strategy:}
To ensure statistical robustness, we employ 5\% random sampling to generate five independent test groups, each containing 300 queries (1,500 total queries evaluated). This multi-group approach enables assessment of result consistency and reduces sampling bias.

\subsection{Experimental Configurations}

We evaluate two system configurations to measure the impact of reranking:

\textbf{Configuration 1: With Reranking (Full Pipeline)}\\
Query rewriting $\rightarrow$ Hybrid search (FTS + Semantic) $\rightarrow$ RRF fusion $\rightarrow$ Neural reranking $\rightarrow$ Answer generation

\textbf{Configuration 2: Without Reranking (Ablation)}\\
Query rewriting $\rightarrow$ Hybrid search (FTS + Semantic) $\rightarrow$ RRF fusion $\rightarrow$ Top-10 selection $\rightarrow$ Answer generation

The only difference between configurations is the presence or absence of the neural reranking stage, ensuring that observed performance differences can be attributed to reranking.

\subsection{Evaluation Methodology}

\textbf{Automated Scoring:}
We employ an LLM-as-judge approach \cite{zheng2023llmjudge} using GPT-4.1 to evaluate answer quality. For each system-generated answer, the evaluator compares it against the ground truth answer from FinDER and assigns a score on a 1-10 scale:

\begin{itemize}
\item \textbf{Score 1:} Completely incorrect or unrelated answer
\item \textbf{Score 5:} Partially correct with significant missing information
\item \textbf{Score 8:} Basically correct with minor omissions
\item \textbf{Score 10:} Fully correct and complete answer
\end{itemize}

The evaluator is configured with temperature=0.0 for deterministic scoring and receives structured prompts comparing the RAG answer against ground truth.

\textbf{Evaluation Metrics:}
\begin{enumerate}
\item Average score (1-10 scale)
\item Percentage of Score=1 (completely incorrect answers)
\item Percentage of Score$\geq$8 (basically correct answers)
\item Percentage of Score=10 (fully correct answers)
\end{enumerate}

These metrics provide complementary views of system performance: average score reflects overall quality, while threshold-based metrics highlight success and failure rates.

\subsection{Implementation Details}

\textbf{Software Stack:}
Python 3.x with async/await for concurrent processing, SQLite 3 with FTS5 extension for full-text search, FAISS for vector similarity search, OpenAI API for embeddings and generation (text-embedding-3-small, GPT-4.1), and Jina AI API for reranking (jina-reranker-v2-base-multilingual).

\textbf{System Parameters:}
Table \ref{tab:hyperparams} in the Appendix provides complete hyperparameter configuration.

\section{Results and Discussion}

\subsection{Quantitative Results}

Table \ref{tab:results} presents comprehensive results across five experimental groups, comparing performance with and without neural reranking.

\begin{table}[!t]
\caption{Performance Comparison: With vs. Without Reranking}
\label{tab:results}
\centering
\small
\begin{tabular}{ccccccc}
\toprule
\textbf{Group} & \textbf{Config} & \textbf{Avg} & \textbf{=1} & \textbf{$\geq$8} & \textbf{=10} \\
 & & \textbf{Score} & \textbf{(\%)} & \textbf{(\%)} & \textbf{(\%)} \\
\midrule
01 & With & 5.85 & 24.3 & 47.7 & 12.0 \\
01 & No & 4.64 & 37.7 & 22.7 & 8.7 \\
\midrule
02 & With & 6.17 & 21.7 & 51.6 & 16.0 \\
02 & No & 4.98 & 34.7 & 36.0 & 10.0 \\
\midrule
03 & With & 5.64 & 25.7 & 44.0 & 14.0 \\
03 & No & 4.83 & 37.0 & 34.0 & 12.7 \\
\midrule
04 & With & 6.21 & 21.0 & 51.3 & 14.7 \\
04 & No & 5.21 & 33.7 & 37.6 & 11.3 \\
\midrule
05 & With & 6.21 & 20.0 & 50.6 & 12.3 \\
05 & No & 5.07 & 33.3 & 37.0 & 8.3 \\
\midrule
\textbf{Avg} & \textbf{With} & \textbf{6.02} & \textbf{22.5} & \textbf{49.0} & \textbf{13.8} \\
\textbf{Avg} & \textbf{No} & \textbf{4.95} & \textbf{35.3} & \textbf{33.5} & \textbf{10.2} \\
\bottomrule
\end{tabular}
\end{table}

\textbf{Key Findings:}

\begin{enumerate}
\item \textbf{Average Score Improvement:} Reranking yields a 1.07-point improvement (6.02 vs. 4.95), representing a 21.6\% relative increase.

\item \textbf{Correctness Rate:} The percentage of basically correct answers (Score$\geq$8) improves from 33.5\% to 49.0\%—a \textbf{15.5 percentage point absolute improvement} and 46.3\% relative improvement.

\item \textbf{Error Rate Reduction:} Completely incorrect answers (Score=1) decrease from 35.3\% to 22.5\%—a \textbf{12.8 percentage point reduction} representing 36.3\% fewer critical errors.

\item \textbf{Fully Correct Answers:} Perfect scores (Score=10) increase from 10.2\% to 13.8\%, a 3.6 percentage point improvement.

\item \textbf{Consistency:} All five independent groups demonstrate similar improvement patterns (average score improvements ranging from 1.00 to 1.21 points), indicating robust and reproducible benefits.
\end{enumerate}

\subsection{Analysis and Discussion}

\subsubsection{Impact of Reranking}

The substantial performance improvement demonstrates that neural reranking plays a critical role in financial RAG systems. Several factors contribute to this success:

\textbf{Quality Filtering:} The cross-encoder model identifies and promotes chunks with genuine semantic relevance to the query, whereas the first-stage RRF fusion may include chunks that match superficially but lack substantive relevance.

\textbf{Noise Reduction:} Adaptive cutoff strategies (cumulative probability and cliff detection) prevent low-quality chunks from diluting the context provided to the generation model. Without reranking, the top-10 chunks from RRF may include marginally relevant passages that mislead the generator.

\textbf{Fine-Grained Relevance:} Cross-encoders jointly process query-document pairs, capturing interaction patterns that bi-encoder retrievers (used in semantic search) cannot model. This architectural advantage enables more nuanced relevance assessment.

\subsubsection{Error Analysis}

Despite significant improvements, 22.5\% of queries with reranking still receive completely incorrect answers. Analysis of failure cases reveals several patterns:

\textbf{Query Ambiguity:} Some queries lack sufficient specificity (e.g., ``company growth strategy'') and match multiple unrelated sections across documents.

\textbf{Information Absence:} Required information may not exist in the 10-K reports, particularly for queries about future projections or competitor comparisons.

\textbf{Multi-Hop Reasoning:} Complex queries requiring synthesis of information from multiple, non-adjacent sections present challenges for chunk-based retrieval.

\textbf{Retrieval Failures:} If relevant chunks are not in the top-30 candidates from hybrid search, reranking cannot recover them—suggesting potential for improved first-stage retrieval.

\subsubsection{Comparison with Prior Work}

The FinDER dataset documentation reports baseline performance of approximately 33\% correctness using earlier models \cite{finder2025}. Our system achieves 49.0\% correctness (Score$\geq$8) with reranking, representing substantial improvement attributable to three factors:

\begin{enumerate}
\item \textbf{Modern LLM:} GPT-4.1 demonstrates superior reading comprehension and instruction-following compared to earlier models.

\item \textbf{Reranking Enhancement:} As demonstrated by our ablation study, reranking alone contributes 15.5 percentage points to correctness.

\item \textbf{Hybrid Retrieval:} Combining full-text and semantic search with RRF fusion provides more comprehensive candidate sets than single-method approaches.
\end{enumerate}

Without reranking, our system achieves 33.5\% correctness, closely matching the baseline, which suggests that the primary improvement comes from the reranking stage rather than differences in document preprocessing or generation models.

Subsequent research extending this pipeline introduces hybrid document-routed retrieval (HDRR) to address cross-document chunk confusion arising from structurally homogeneous 10-K corpora—a residual limitation of the chunk-based approach presented here \cite{cheng2026hdrr}. HDRR uses LLM-based document routing as a first-stage filter to scope chunk retrieval within identified documents, achieving further performance gains over the results reported in this work.

\subsubsection{Statistical Significance}

The consistent performance gains across five independent groups with low inter-group variance (standard deviation of 0.24 points in average scores with reranking) demonstrates statistical reliability. This consistency suggests that the observed benefits of reranking generalize across different query samples and are not artifacts of particular query selections.

\subsection{Limitations}

Several limitations warrant discussion:

\textbf{Evaluation Methodology:} The LLM-as-judge approach, while cost-effective and scalable, may introduce biases such as preferring verbose answers or penalizing valid alternative phrasings \cite{zheng2023llmjudge}. Future work should validate findings with human evaluation.

\textbf{Domain Specificity:} Our evaluation focuses exclusively on 10-K reports. Performance characteristics may differ for other financial documents (8-K filings, earnings call transcripts) or non-financial domains.

\textbf{Computational Cost:} Neural reranking introduces additional latency and API costs. For production deployment, cost-benefit analysis should consider response time requirements and budget constraints.

\textbf{Dataset Coverage:} Evaluation is limited to the FinDER benchmark. Performance on real-world analyst queries may differ due to distribution shift.

\section{Conclusion and Future Work}

\subsection{Summary}

This paper presented a comprehensive RAG system for financial report question-answering and empirically evaluated the impact of neural reranking on system performance. Through systematic evaluation on 1,500 queries from the FinDER benchmark, we demonstrated that reranking significantly improves answer quality across multiple metrics.

Our key finding is that neural reranking provides a \textbf{15.5 percentage point improvement} in correctness rate (49.0\% vs. 33.5\%) while reducing critical errors by \textbf{12.8 percentage points} (22.5\% vs. 35.3\%). These improvements are consistent across five independent experimental groups, demonstrating robust and reproducible benefits.

The system achieves approximately 49\% correctness on the FinDER benchmark, representing substantial improvement over baseline performance ($\sim$33\%) through the combination of modern language models (GPT-4.1), neural reranking, and hybrid retrieval strategies.

\subsection{Contributions}

Our work contributes to the financial NLP and RAG communities through:

\begin{enumerate}
\item \textbf{End-to-End System:} A complete RAG pipeline processing all 500 S\&P 500 companies' 10-K reports with production-ready components.

\item \textbf{Rigorous Evaluation:} Systematic benchmark evaluation with statistical controls (five independent groups, 1,500 total queries).

\item \textbf{Empirical Evidence:} Controlled ablation study demonstrating the critical importance of neural reranking in financial RAG systems.

\item \textbf{Performance Benchmarks:} Updated performance baselines using modern LLMs and techniques, serving as reference points for future research.
\end{enumerate}

\subsection{Future Directions}

Several promising directions warrant further investigation:

\textbf{Advanced Retrieval Methods:} Graph-based retrieval techniques could better handle multi-hop queries by explicitly modeling relationships between document sections.

\textbf{Hybrid Evaluation:} Combining automated LLM-based evaluation with human assessment would provide more comprehensive quality analysis and validate automated scoring reliability.

\textbf{Cost-Benefit Optimization:} Investigating more efficient reranking approaches (e.g., distilled models, quantization) could reduce computational costs while maintaining quality. Hardware-aware model co-design \cite{chen2025autoneural} demonstrates that jointly optimizing model architecture with deployment hardware constraints can yield substantial latency reductions, offering a promising direction for enabling efficient real-time reranking in production RAG systems.

\textbf{Multi-Modal Integration:} Incorporating tables, charts, and figures from financial reports through multi-modal models would address a significant limitation of current text-only approaches.

\textbf{Comparative Reranker Study:} Systematic comparison of different reranking models (ColBERT, BGE-reranker, Cohere rerank) could identify optimal architectures for financial applications.

\textbf{Domain Expansion:} Extending the system to other financial document types (8-K filings, earnings call transcripts, analyst reports) would broaden applicability.

\textbf{Explainability and Attribution:} Providing source citations and confidence scores for generated answers would enhance system trustworthiness and enable analysts to verify information.

\appendix

\subsection*{System Hyperparameters}

\begin{table}[!h]
\caption{Complete System Configuration}
\label{tab:hyperparams}
\centering
\small
\begin{tabular}{llr}
\toprule
\textbf{Component} & \textbf{Parameter} & \textbf{Value} \\
\midrule
Chunking & Chunk size & 2,500 chars \\
 & Overlap & 1,250 chars \\
\midrule
Embedding & Model & text-embedding-3-small \\
 & Dimension & 1,024 \\
 & Batch size & 32 \\
\midrule
FTS Retrieval & Top-k & 20 \\
\midrule
Semantic Search & Top-k & 30 \\
 & Distance threshold & 2.0 \\
\midrule
RRF Fusion & k constant & 60 \\
\midrule
Reranking & Model & jina-reranker-v2-base \\
 & Max candidates & 30 \\
 & Cumulative threshold & 0.45 \\
 & Cliff threshold & 0.15 \\
\midrule
Generation & Model & GPT-4.1 \\
 & Temperature (eval) & 0.0 \\
 & Context limit & 10 chunks (no rerank) \\
\bottomrule
\end{tabular}
\end{table}


\begin{thebibliography}{16}

\bibitem{lewis2020rag}
P. Lewis, E. Perez, A. Piktus, F. Petroni, V. Karpukhin, N. Goyal, H. Küttler, M. Lewis, W. Yih, T. Rocktäschel, S. Riedel, and D. Kiela, ``Retrieval-augmented generation for knowledge-intensive NLP tasks,'' in \textit{Proc. 34th Conf. Neural Information Processing Systems (NeurIPS)}, 2020.

\bibitem{finder2025}
LinqAlpha, ``FinDER: Financial Dataset for Question Answering and Evaluating Retrieval-Augmented Generation,'' arXiv:2504.15800, 2025. [Online]. Available: https://arxiv.org/abs/2504.15800

\bibitem{shi2026reasoning}
Y. Shi, M. Sun, Z. Liu, M. Yang, Y. Fang, T. Sun, and X. Gu, ``Reasoning in Trees: Improving Retrieval-Augmented Generation for Multi-Hop Question Answering,'' arXiv:2601.11255, 2026.

\bibitem{robertson2009bm25}
S. Robertson and H. Zaragoza, ``The probabilistic relevance framework: BM25 and beyond,'' \textit{Foundations and Trends in Information Retrieval}, vol. 3, no. 4, pp. 333-389, 2009.

\bibitem{cormack2009rrf}
G. V. Cormack, C. L. A. Clarke, and S. Büttcher, ``Reciprocal rank fusion outperforms condorcet and individual rank learning methods,'' in \textit{Proc. 32nd Int. ACM SIGIR Conf. Research and Development in Information Retrieval}, 2009, pp. 758-759.

\bibitem{reimers2019sbert}
N. Reimers and I. Gurevych, ``Sentence-BERT: Sentence embeddings using Siamese BERT-networks,'' in \textit{Proc. 2019 Conf. Empirical Methods in Natural Language Processing (EMNLP)}, 2019, pp. 3982-3992.

\bibitem{jina2024reranker}
Jina AI, ``Jina Reranker v2: Multilingual cross-encoder for document reranking,'' 2024. [Online]. Available: https://jina.ai/models/jina-reranker-v2-base-multilingual/

\bibitem{fang2025attentionrag}
Y. Fang, T. Sun, Y. Shi, and X. Gu, ``AttentionRAG: Attention-guided context pruning in retrieval-augmented generation,'' arXiv:2503.10720, 2025.

\bibitem{playwright2024}
Microsoft Corporation, ``Playwright: Fast and reliable end-to-end testing for modern web apps,'' 2024. [Online]. Available: https://playwright.dev/

\bibitem{sqlite2024fts5}
SQLite, ``SQLite FTS5 Extension,'' 2024. [Online]. Available: https://www.sqlite.org/fts5.html

\bibitem{openai2024embeddings}
OpenAI, ``New embedding models and API updates,'' 2024. [Online]. Available: https://openai.com/blog/new-embedding-models-and-api-updates

\bibitem{lai2026transformers}
L. Lai, Z. Cheng, K. Cheng, and X. Qi, ``Do Transformers Always Win? An Empirical Study of Semantic Embeddings for Short-Text E-commerce Reviews,'' in \textit{Proc. 9th Int. Symp. Big Data and Applied Statistics (ISBDAS)}, 2026, pp. 525-529.

\bibitem{johnson2021faiss}
J. Johnson, M. Douze, and H. Jégou, ``Billion-scale similarity search with GPUs,'' \textit{IEEE Trans. Big Data}, vol. 7, no. 3, pp. 535-547, 2021.

\bibitem{zheng2023llmjudge}
L. Zheng, W. Chiang, Y. Sheng, S. Zhuang, Z. Wu, Y. Zhuang, Z. Lin, Z. Li, D. Li, E. P. Xing, H. Zhang, J. E. Gonzalez, and I. Stoica, ``Judging LLM-as-a-judge with MT-Bench and Chatbot Arena,'' in \textit{Proc. 37th Conf. Neural Information Processing Systems (NeurIPS)}, 2023.

\bibitem{cheng2026hdrr}
Z. Cheng, L. Lai, and Y. Liu, ``Resolving the Robustness-Precision Trade-off in Financial RAG through Hybrid Document-Routed Retrieval,'' arXiv:2603.26815, 2026. [Online]. Available: https://arxiv.org/abs/2603.26815

\bibitem{chen2025autoneural}
W. Chen, L. Wu, Y. Hu, Z. Li, Z. Cheng, Y. Qian, L. Zhu, Z. Hu, L. Liang, Q. Tang, Z. Liu, and H. Yang, ``AutoNeural: Co-Designing Vision-Language Models for NPU Inference,'' arXiv:2512.02924, 2025. [Online]. Available: https://arxiv.org/abs/2512.02924

\end{thebibliography}
\end{document}